\documentclass[letterpaper, 10 pt, conference]{ieeeconf}  

\IEEEoverridecommandlockouts                              

\usepackage{math}
\usepackage{amsmath}

\title{\LARGE \bf
PoCo: Point Context Cluster for RGBD Indoor Place Recognition}

\author{Jing Liang$^{1}$, Zhuo Deng$^{2}$, Zheming Zhou$^{2}$, Omid Ghasemalizadeh$^{2}$, Dinesh Manocha$^{1}$, \\
Min Sun$^{2}$, Cheng-Hao Kuo$^{2}$, Arnie Sen$^{2}$
\thanks{$^{1}$ University of Maryland, College Park; $^{2}$ Amazon, Bellevue, WA, USA; }%
\thanks{Code: \url{https://github.com/jingGM/PoCo-CCR}}
}

\newcommand{\rev}[1]{{\color{black}#1}}

\begin{document}

\maketitle
\thispagestyle{empty}
\pagestyle{empty}

\begin{abstract}
We present a novel end-to-end algorithm (PoCo) for the indoor RGB-D place recognition task, aimed at identifying the most likely match for a given query frame within a reference database. The task presents inherent challenges attributed to the constrained field of view and limited range of perception sensors. We propose a new network architecture, which generalizes the recent Context of Clusters (CoCs) to extract global descriptors directly from the noisy point clouds through end-to-end learning. Moreover, we develop the architecture by integrating both color and geometric modalities into the point features to enhance the global descriptor representation. We conducted evaluations on public datasets ScanNet-PR and ARKit with 807 and 5047 scenarios, respectively. PoCo achieves SOTA performance: on ScanNet-PR, we achieve R@1 of $64.63\%$, a $5.7\%$ improvement from the best-published result CGis ($61.12\%$); on Arkit, we achieve R@1 of $45.12\%$, a $13.3\%$ improvement from the best-published result CGis ($39.82\%$). In addition, PoCo shows higher efficiency than CGis in inference time ($1.75$X-faster), and we demonstrate the effectiveness of PoCo in recognizing places within a real-world laboratory environment. Video: \url{https://youtu.be/D8dObAeMiCw}; 
\end{abstract}

\section{Introduction}
For mobile robots, localization is a critical capability that enables them to determine their position within a known environment~\cite{ijcai2021p603}. Visual Place Recognition (VPR) is typically formulated as an image retrieval problem, i.e., retrieving candidate frames from a database by comparing similarities to a given query frame~\cite{lowry2015visual, ijcai2021p603}. Consequently, the design of an accurate place recognition system plays an important role in localization tasks by matching current observations with previously visited scenarios from the database~\cite{zhu2023r2former, ming2022cgis, arandjelovic2016netvlad}. It finds extensive use in applications like  navigation~\cite{stumm2013probabilistic, mirowski2018learning}, and loop closure in SLAM~\cite{lowry2015visual} etc.
However, the place recognition problem is not trivial. There could be various challenging environments~\cite{lowry2015visual,ijcai2021p603, liu2023survey, zhu2023r2former,ming2022cgis} with illumination changes, objects changes, dynamic objects, etc. Furthermore, different sensors could be employed in specific environmental conditions. For example, Lidar~\cite{uy2018pointnetvlad} and Mono-camera~\cite{zhu2023r2former,lowry2015visual} are used for outdoor place recognition, while RGB-D cameras~\cite{ming2022cgis, yudin2022hpointloc} are widely used for indoor scenarios. How to effectively process the perception data from the sensor is also challenging~\cite{qi2017pointnet, cocs, wu2023pointconvformer}. In this paper, our main focus is on addressing the problem of indoor RGB-D point cloud-based place recognition.

\textbf{Challenges for Indoor Place Recognition:} While indoor and outdoor place recognition encounter similar challenges~\cite{liu2023survey, lowry2015visual}, such as varying illumination conditions, dynamic environments, and changes in view perspective, indoor place recognition possesses distinct characteristics.
Firstly, indoor environments often exhibit less variability which makes it challenging to distinguish between different indoor spaces that may appear visually similar. Secondly, indoor perceptions often exhibit shorter ranges compared to those of outdoor scenes~\cite{ming2022cgis}. 
Finally, objects in indoor scenes are typically closer to sensors compared to those in outdoor scenes, even a slight camera motion can result in significant visual appearance changes within the field of view. Consequently, approaches tailored for outdoor scenarios may not translate effectively to indoor tasks.

\begin{figure}
    \centering    
    \includegraphics[width=\linewidth]{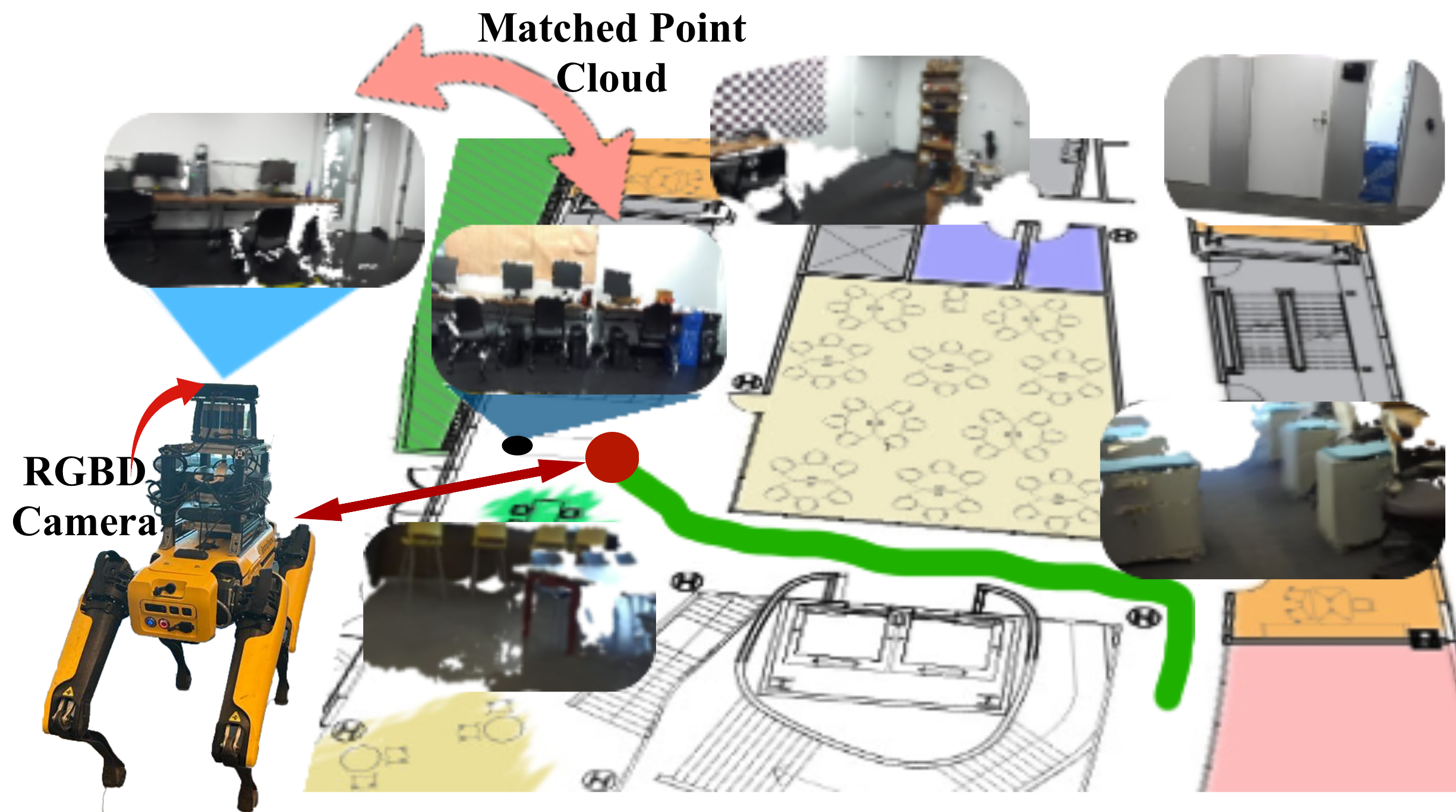}
    \caption{As shown in the figure, the database contains several frames captured by RGB-D cameras in different rooms. The robot moves along the green trajectory and performs place recognition in real time to locate itself. When the robot moves to the red circle, it observes the blue frame and successfully finds the best-matched frame in the database.}
    \label{fig:front_page}
    \vspace{-1.5em}
\end{figure}

\textbf{RGB-D Place Recognition is Under Development:} A common approach for place recognition involves extracting global descriptors from both query and database frames, followed by ranking retrieved candidate frames based on the computed similarities of these global descriptors~\cite{lowry2015visual, ming2022cgis, yudin2022hpointloc}. Many approaches have been proposed for RGB place recognition~\cite{arandjelovic2016netvlad, zhu2023r2former, lowry2015visual}, which are mostly for outdoor place recognition because RGB camera captures rich, long-range and dense color information. For indoor scenarios, other than the RGB camera, the RGB-D camera shows more capability of perceiving dense depth images, whereas the geometric information should also be considered for the place recognition task. PointNet-VLAD~\cite{uy2018pointnetvlad}, MinkLoc-3D~\cite{komorowski2021minkloc3d} and Indoor DH3D~\cite{du2020dh3d, yang2022fd} process point clouds for place recognition, but they are not designed to process color information. To utilize both color and geometric information, Sizikova et al.~\cite{inbook} process RGB and depth images separately by convolutional layers and concatenate them as a joint descriptor. To completely fuse the geometric and color information in the global descriptors, CGiS-Net~\cite{ming2022cgis} utilizes KP-Conv~\cite{thomas2019kpconv} to process both color and geometric data. In our approach, we present a novel structure to improve the efficacy of processing geometric information and also propose a better feature encoder for RGB-D place recognition tasks.

\textbf{Better Feature Extraction Can Be Used:} Because place recognition (retrieval) problem requires encoding a global descriptor for each frame, feature extraction plays an important role in the process. Convolution~\cite{thomas2019kpconv, uy2018pointnetvlad, arandjelovic2016netvlad} and Tranformers~\cite{zhu2023r2former, dosovitskiy2021an} are used for feature processing in place recognition tasks. Vision transformers (ViTs)~\cite{dosovitskiy2020image} show better performance in the feature extraction for different vision tasks~\cite{deininger2022comparative}. CoCs~\cite{cocs} is recently introduced to efficiently extract features from RGB images with less computational cost but with comparable performance in vision tasks as ViTs. CoCs leverage learned centers with larger receptive fields in an image to cluster and aggregate local pixels. Then it dispatches the center features to local features for further enhancement. However, CoCs is only designed for 2D images, and cannot be directly applied for point cloud inference. Inspired by the novel aggregation and dispatch mechanism where semantic-related features will be dynamically grouped and learned together, generalize this idea to jointly extract appearance and geometric features from RGB-D data for place recognition.


\textbf{Main Results:} We propose a novel architecture, as shown in Fig. \ref{fig:front_page}, to handle both color appearance and geometric features from the RGB-D point cloud. We generalize CoCs to process point clouds and make it capable of extracting both color appearance and geometric features jointly from RGB-D data. 
Our approach is trained and evaluated in the ScanNet-PR~\cite{ming2022cgis} and ARKit~\cite{arkit} datasets with 807 and 5047 indoor scenarios, respectively. These two datasets exhibit various illuminations, layouts, sizes, and color features across these scenarios. In particular, the ARKit dataset is collected by a mobile device, resulting in lower depth resolution and sparser RGB-D point clouds compared to ScanNet-PR. Our contributions include:

\begin{enumerate}
    \item We propose a novel end-to-end architecture to generalize CoCs concept from operating on 2D image domain to point clouds, where local point features are learned by interacting with all higher-level center points in a novel aggregation-and-dispatch approach.
    \item We develop the architecture to jointly process different modalities, color and geometric information, to improve the performance in indoor RGB-D place recognition task. Especially, we explicitly encode geometric information into point features to enhance the representation of global descriptors.
    \item Our method consistently outperforms SOTA baseline models by a significant margin on multiple datasets. We observe a relative improvement of $5.7\%$ and $13\%$ in Recall@1 over baselines on challenging large-scale datasets, ScanNet-PR~\cite{ming2022cgis} and ARKit ~\cite{arkit}.
\end{enumerate}

\section{Related Works}

\textbf{Visual Place Recognition:} Place recognition is often formulated as an image retrieval problem, where the candidate frames are ranked by calculating similarities between their global descriptors and the descriptor of a query frame~\cite{ming2022cgis, zhu2023r2former}.  Traditional place recognition methods extract classical RGB image features, SIFT, SURF,  etc~\cite{david2004distinctive, cummins2008fab,galvez2012bags} and use the features to compose descriptors, e.g. Bag-of-Words~\cite{sivic2008efficient}, to match the frames. Then learning-based approaches generate more representative and emperical features for visual place recognition and improved the performance~\cite{arandjelovic2016netvlad, lowry2015visual}. Especially, CNN-based methods~\cite{uy2018pointnetvlad, jin2017learned, gordo2017end} become predominant methods for visual place recognition tasks. By extending feature-engineered VLAD into a data-driven learning approach, NetVLAD-based approaches~\cite{arandjelovic2016netvlad} have resulted in higher accuracy in terms of matching than the traditional methods. However, extracting 3D spatial geometric features from static images is still challenging. On the other hand, numerous works directly consume point clouds as their inputs. ~\cite{uy2018pointnetvlad} employs PointNet~\cite{qi2017pointnet} structure and upgrade ~\cite{arandjelovic2016netvlad} to directly extract features from a 3d point cloud. To improve the limitation of ~\cite{uy2018pointnetvlad} on capturing local geometric structures, ~\cite{komorowski2021minkloc3d} proposed to compute more discriminative 3D descriptors based on a sparse voxelized point cloud representation and sparse 3D convolutions. ~\cite{du2020dh3d, yang2022fd} designed a Siamese network that jointly learns 3D local feature detection and description directly from raw 3D points. While point clouds offer numerous advantages such as encoded 3D spatial information and viewpoint invariance, it is arguable that integrating 3D features with 2D color features into the descriptors could potentially enhance performance further as this integration exploits the complementary nature of 3D and 2D features, leveraging the strengths of both modalities for more comprehensive scene representation.

\textbf{Indoor RGB-D Place Recognition:} This research field is under-explored as only a few works are available for indoor RGB-D place recognition. For example, ~\cite{yudin2022hpointloc} employs Patch-NetVLAD~\cite{hausler2021patch} for image retrieval and the estimate 6 DoF camera poses based on refined feature matches with synthetic RGB-D images. The system combines various components from existing techniques and follows a piecewise optimization approach, whereas ours adopts an end-to-end learning approach. The most relevant to our work is CGiS-Net~\cite{ming2022cgis}, which aggregates both color, semantic, and geometric information together using real RGB-D images, and its geometry feature is extracted by KPConv~\cite{thomas2019kpconv}. Again, CGis-Net has two distinct training stages for semantic encoder/decoder and feature encoder respectively, whereas ours is end-to-end and the network architecture is significantly different from theirs. 

\textbf{Feature Processing:} Traditional methods~\cite{rublee2011orb, cummins2008fab} use manual-crated features for place recognition. After that CNN-based methods demonstrated better performance in different vision tasks~\cite{uy2018pointnetvlad, jin2017learned, gordo2017end} and also place recognition. NetVLAD-based approaches~\cite{arandjelovic2016netvlad} show good performance for RGB place recognition, and KPConv-based approaches~\cite{Shi2023LiDARBasedPR, ming2022cgis} show promising results in point cloud place recognition. Vision transformers (ViTs)~\cite{dosovitskiy2020image} is the next generation of the feature extractor and outperform CNN-based approaches in multiple vision tasks~\cite{touvron2021training, berton2022deep, zhu2023r2former}. Recently, a new light-weighted feature extraction method Context of Clusters (CoCs)~\cite{cocs}, is proposed and demonstrates state-of-the-art performance in different vision tasks but is more computationally effective than ViTs. However, CoCs is only designed for RGB features. In our work, we generalize it to RGB-D point cloud feature extraction. 

\section{Our Approach}
In this section, we first formulate the problem in Section \ref{sec:problem_formulation}, then describe the architecture of our PoCo methon. Finally, we discuss the training strategies in Section \ref{sec:training}.

\subsection{Problem Formulation}
\label{sec:problem_formulation}

Following a similar paradigm to VPR tasks~\cite{arandjelovic2016netvlad, zhu2023r2former}, we define the indoor RGB-D place recognition task as a point cloud retrieval problem. In subsequent sections, we will refer to each frame as a colorized point cloud. Denote a query frame $\Q \in \cq$ and a candidate frame  $\D\in\cd$, where $\cq$ and $\cd$ represent query set and database respectively. In general, the model $\mathcal{M}$ transforms each frame into a global descriptor via representation learning, $\mathcal{M}: frame \rightarrow \mathbf{v} \in \mathbb{R}^n$. The goal is to retrieve likely matched candidate frames from the database based on descriptor similarities between query and database. Therefore, the model is expected to learn a representation such that positive frame pairs are close together in the embedding space, while negative frame pairs are far apart. To be more specific, $s(\mathcal{M}(\Q), \mathcal{M}(\D_p)) \gg s(\mathcal{M}(\Q), \mathcal{M}(\D_n))$ where $s(\cdot)$ is the similarity function, and $\D_p$ and $\D_n$ are positive and negative frames.

\begin{figure*}[h!]
    \centering
    \includegraphics[width=0.9\linewidth]{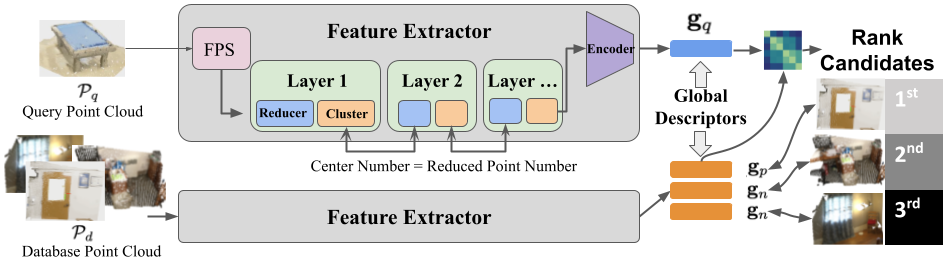}
    \caption{\textbf{PoCo Architecture.} The input contains query and database frames and the model generates a global descriptor for each frame, and the descriptors are used to rank database frames by the similarities to the query frame. Note that the extractor consists of Farthest Point Sampling (FPS), an encoder, and layers of Reducers and Cluster blocks (see Fig.~\ref{fig:reducer},~\ref{fig:cocs}). }
    \label{fig:architecture}
    \vspace{-2em}
\end{figure*}

\subsection{Architecture of Our PoCo Model}
\label{sec:architecture}

The architecture of our PoCo model is shown in Fig. \ref{fig:architecture}. The input frames possess identical tensor shapes as $N\times 9$, where $N$ is the number of points in a frame and each point is asociated with a 9-D feature vector, i.e., color $\set{r, g, b}$, position $\set{x, y, z}$ and normal $\set{n_x, n_y, n_z}$. As the Context of Clusters (CoCs)~\cite{cocs} is originally designed for 2D images, PoCo generalizes it to process 3D point cloud. In the high-level, the basic layer of the network is consisting of one Point Reducer block and one Context Cluster block. The Point Reducer downsamples point cloud and aggregate point features, while the Context Cluster enhances the point features further in an aggregation and dispatch way. To facilitate jointly learning of color and geometric features, each point feature vector is designed to have two parts: feature part and geometry part. The feature part carries the learned embedding, while geometry part stores fixed content, i.e., position and normal. The motivation to include the normal is that helps to encode points relative positional relationships for better model generalizability.

\begin{figure}
    \centering
    \includegraphics[width=0.8\linewidth]{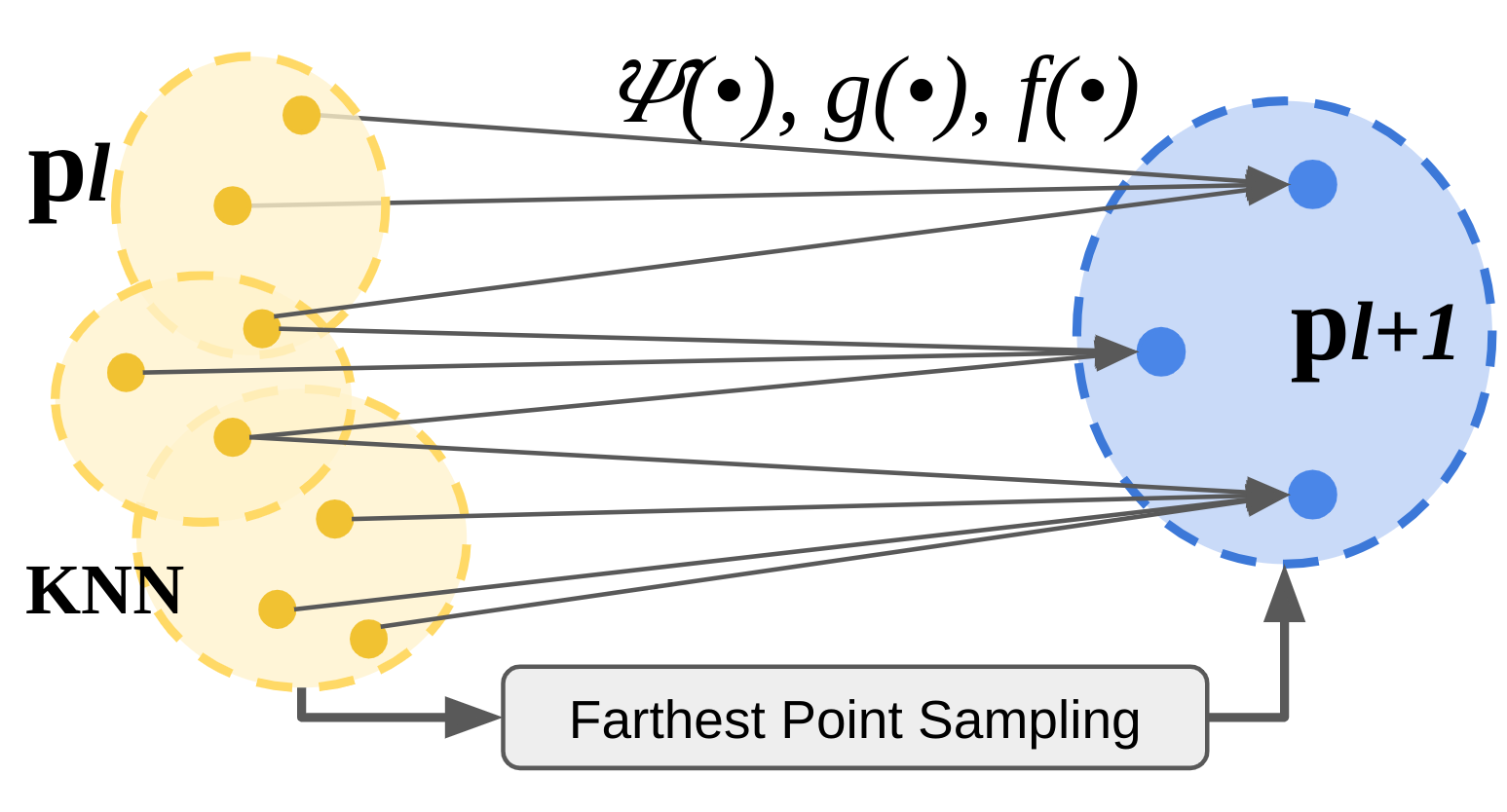}
    \caption{\textbf{Reducer Block.} The blue circles are points in different levels. In Reducer Block, we use the Farthest Point Sampling method to downsample points $\P^l$ to $\P^{l+1}$ and aggregate the geometric and feature information of the K-nearest neighbors to the downsampled $\P^{l+1}$ by Eq.~\ref{eq:pointconvformer}.}
    \label{fig:reducer}
    \vspace{-1em}
\end{figure}

\textbf{Reducer Blocks:} Each reducer block is used to (1) downsample points population from the last layer; (2) estimate relative geometric relationships and aggregate embedding features for reduced points.  For points downsampling, we choose the Farthest Point Sampling (FPS) strategy~\cite{eldar1997farthest} for its excellent computational efficiency, \rev{with the complexity of $O(nk)$}, while preserving the structure of the point cloud. Then for each point $p\in \P^{l+1}_r$, we apply KNN~\cite{kramer2013k} to choose the K nearest points from $\P^{l}_r$ as shown in Fig. \ref{fig:reducer}, where $\P^{l}_r$ and $\P^{l+1}_r$ represent input and output point sets in the reducer block. The feature aggregation from $\P^l_r$ to $\P^{l+1}_r$ is based on PointConvFormer~\cite{wu2023pointconvformer} as shown in Eq.~\ref{eq:pointconvformer}, where $\cn(p)$ represents $p$'s neighbor set. $f_{1,2,3,4}(\cdot)$ are learnable Linear Layers, and $\psi(\cdot)$ is a multi-head attention module calculating the similarity score between $f(p_k)$ and $f(p)$.

\begin{align}
    \f_p &= \sum_{p_k\in \cn(p)} f_4(g(p_k,p))\psi(f_1(p_k), f_2(p))f_3(p_k).
    \label{eq:pointconvformer}
\end{align}
The vector $g(p_k, p)$ encodes coordinate-independent geometric information as shown in Eq.\ref{eq:geos}, where $\r_k=p-p_k$ is a translation vector from $p_k$ to $p$, $\n$ and $\n_k$ are normals of $p$ and $p_k$. Thus, $g(p_k, p)$ only encodes the relative geometric information between the two points~\cite{li2021devils}.
\begin{align}
    \hat{\r}_k = \frac{\r_k}{\norm{\r_k}}; \;\; &\v = \frac{\n_k-(\n_k\cdot \hat{\r}_k)\hat{\r}_k}{\norm{\n_k-(\n_k\cdot \hat{\r}_k)\hat{\r}_k}}; \;\; \w=\frac{\hat{\r}_k\times \v}{\norm{\hat{\r}_k\times \v}};\nonumber
\end{align}
 \begin{align}
    g(p_k, p) = &[\n\cdot \n_k, \frac{\r_k\cdot \n_k}{\norm{\r_k}}, \frac{\r_k\cdot \n}{\norm{\r_k}}, \n\cdot\v, \n\cdot\w, \r_k\cdot \n_k, \nonumber\\
    &\r_k\cdot(\n\times\n_k), \norm{\r_k}], 
    \label{eq:geos}
\end{align}

\begin{figure}
    \centering
    \includegraphics[width=\linewidth]{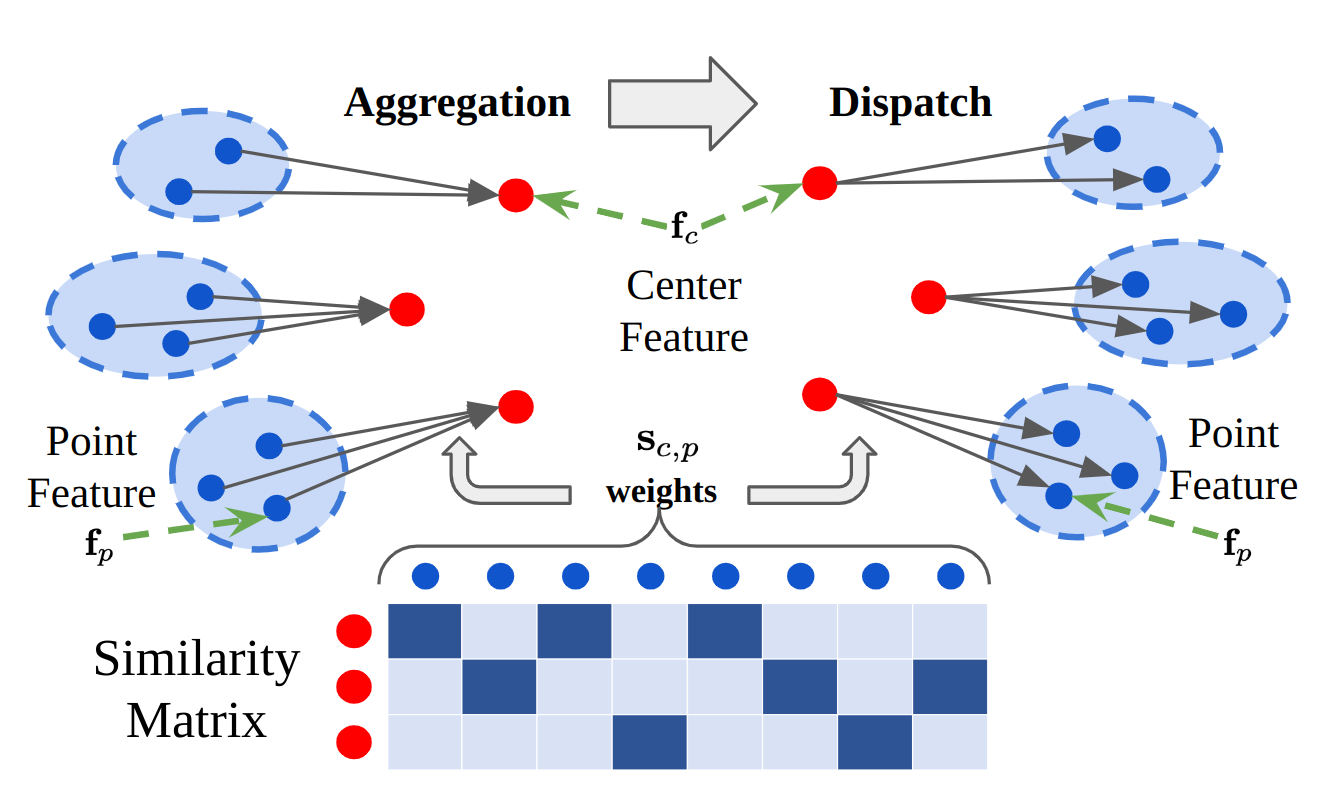}
    \caption{\textbf{Cluster Block.} Red points are centers downsampled from the blue points. The blue points are grouped by the similarity of w.r.t. the red centers. Then the features of the blue points are aggregated to the centers by Eq.~\ref{eq:cluster} and then the center features are dispatched to the points by Eq.~\ref{eq:dispatch}.}
    \label{fig:cocs}
    \vspace{-2em}
\end{figure}
\textbf{Context Cluster Blocks:} Points in the block are dynamically clustered into groups/centers based on their learned affinities in the embedding space during the aggregation stage. As semantically correlated point features are more likely to be grouped together, each center point can effectively learn richer features from its associated member points. After that, aggragated center features are adaptively dispatched to each member point based on the similarity. In this way, member points implicitly communicate with each other via their center point and learn to optimize features jointly. We provide an overview of context cluster in Fig.\ref{fig:cocs} and will delve into further details in the following sections.

\textbf{Aggregation:} For the point set $\P^{l+1}_r$, we downsample it to generate center points $\P^{l+1}_c$, where the cardinality $\abs{\P^{l+1}_r} = N_r > \abs{\P^{l+1}_c} = N_c$. \rev{$N_c$ is chosen by FPS, less than half of the $N_r$.} \rev{Each initialized center feature $\f'_c$ is calculated} as the mean vector of its K spatially nearest point features $\{\f_p\}$ from $\P^{l+1}_r$. As is shown in Fig.~\ref{fig:cocs},in the aggregation stage, we firstly calculate the cosine similarity scores between all the centers and all the points: $s_{c,p} = \sigma(\alpha \cdot cosine(\f_p, \f'_c)+\beta)$, where $\sigma(\cdot)$ is the sigmoid activation function and $\alpha$ and $\beta$ are trainable parameters. Then point features are aggregated into center points according to previous computed similarities. The center feature $\f'_c$ is incorporated as an anchor feature for numerical stability as well as further emphasize the locality in Eq. \ref{eq:cluster}
\begin{align}
    \f_c = \frac{1}{\cc} \left( \f'_c + \sum_{p=1}^{N_r}  \s_{c,p} \cdot \f_p \right),
    \label{eq:cluster}
\end{align}
$\cc=1+\sum_{p=1}^N \s_{c,p}$, is the normalization factor.

\textbf{Dispatch:} The previous learned similarities are used to dispatch the aggregated center features to their member cluster points via Eq.\ref{eq:dispatch} and $h(\cdot)$ indicates Linear Layers:
\begin{align}
    \f_p = \f_p + h(s_{c,p} \cdot \f_c )).
    \label{eq:dispatch}
\end{align}

\textbf{Global Descriptor Extraction:} The Encoder shown in Fig. \ref{fig:architecture} aggregates all the point features from the last Context Cluster block into a global descriptor, which is used to measure the similarity between the query frame and candidate frames of database. We adopt cosine similarity as the metric function. The encoder is implemented as a variant of the Reducer Block, which is configured to have a single point as the output. Its input point features are all normalized and aggregated by the geometric and feature weights as described in Equation \ref{eq:pointconvformer}. In the experiments, the dimension of the global descriptor is set to 256. 

\subsection{Training}
\label{sec:training}

During training, models are optimized to distinguish positive examples from negative examples. We utilize the widely used Triplet Loss~\cite{schroff2015facenet} for this purpose. In addition, we also aim to provide guidance to estimate the similarity within certain boundaries for stable training. Therefore, Circle Loss is also employed in the training loss~\cite{sun2020circle}.

The circle loss, $\cl_c$, is defined as Equation~\ref{eq:circle}. $m=0.2$ is a margin to reinforce the optimization \rev{and $\delta_p = 1 - m$ and $\delta_n = m$}. $cl(\cdot)$ is a clamp function, which clamps the value to 0 if it is negative.  $\gamma=1$ is a hyperparameter. $s_p$ and $s_n$ are cosine similarities of  the query-positive frame pair and query-negative frame pair, respectively. The circle loss guides the similarities between the query and positive cases to 1 and negative cases to 0.
\begin{align}
    a_p = cl(s_p-1-m, 0), a_n = cl(s_n+m, 0) \\
    \cl_c = softplus(\log \sum \exp(\gamma a_n (s_n - \delta_n)) - \notag \\
    \log \sum \exp(\gamma a_p (s_p - \delta_p)))
    \label{eq:circle}
\end{align}

Triplet loss, $\cl_t$, is defined as Equation \ref{eq:triplet}. Triplet loss is to maximize the difference between the distances of the query-positive frame pair and query-negative frame pair. \rev{The distance between the query and database descriptors is calculated by the L2 distance function, and the similarity in the circle loss is cosine similarity. Those two metrics can be converted by $d(\x,\y)= \norm{\x -\y}^2_2 = 2-2 \;cos(\x,\y)$ for two vectors $\set{\x,\y}$.} The margin is set as $m=0.2$.

\begin{align}
    \cl_t = \max(d(\g_p, \g_q) - d(\g_n, \g_q) + m, 0)
    \label{eq:triplet}
\end{align}

In the final loss function, we combine the two losses $\cl = \alpha \cl_c + \beta \cl_t$. In experiments, we find the best weights for these two terms as $\alpha=10$ and $\beta=0.1$ via hyper-parameters tuning. Triplet loss can rapidly converge, facilitating the differentiation between positive and negative examples. And circle loss aids in refining the estimation process, resulting in more accurate similarity values that closely align with the boundaries.

\section{Experiments}
\label{sec:exp}
In this section, we firstly introduce two large-scale datasets with various challenging scenarios for training and evaluation. Then we describe the settings of the implementation. For evaluation, we demonstrate our approach outperforms other state-of-the-art indoor RGB-D based or point-cloud based approaches by both quantitative and qualitative results. Then we conduct ablation studies to evaluate contribution and impact of individual components on the indoor RGBD place recognition task.

\subsection{Dataset and Training Settings}

Our approach is trained and evaluated on two indoor RGB-D datasets, ScanNet-PR~\cite{ming2022cgis} and ARKit~\cite{arkit}. They are popular large scale datasets suitable for data-driven learning approaches and encompass a variety of object and scene categories presenting diverse illumination conditions. ScanNet-PR is derived from the ScanNet dataset~\cite{dai2017scannet} captured by a commodity RGB-D sensor combined with a iPad RGB camera and the ARKit dataset is captured by an iPad camera together with a dense Lidar scanner. 
The datasets contain numerous scenarios, each representing a room with multiple frames. Each frame contains a point cloud with both color ($\set{r, g, b}$) and geometric ($\set{x, y, z}$) information for each point. However, the number of points varies for different frames. Following previous work dataset split settings,  we split ScanNet-PR's 807 scenarios into training (565), validation (142), and testing (100) datasets. For the ARKit dataset, we employ a similar processing strategy as that used for ScanNet-PR~\cite{ming2022cgis}. The ARKit dataset has 5047 scenarios and we split it into training (3958), validation (1089), and testing (100) datasets. 

Our PoCo model is implemented by Pytorch and trained on 8 Tesla-V100 GPUs. The input points are constrained to 2000 points through voxelization. Besides $\set{r, g, b}$ and $\set{x, y, z}$, we also use the normal vector $\set{n_x, n_y, n_z}$ of each point to provide additional relative geometric information as in Eq.~\ref{eq:geos}, which is estimated by the plane within a 0.2-meter radius of the point. For the training strategy, we use cosine annealing scheduler~\cite{loshchilov2016sgdr} with Adam optimizer~\cite{kingma2014adam}. The learning rate is from $10^{-4}$ to $10^{-7}$.

\subsection{Implementations of Comparison and Ablation Study}
To ensure a fair comparison, we use the SOTA evaluation strategy of CGiS-Net~\cite{ming2022cgis}, which is the current best approach in the ScanNet-PR dataset. In each scenario of the dataset, the database frames are chosen by a distance threshold of 3 meters, and other frames are query frames. For each query frame, we use the Recall metric to evaluate the place recognition performance: The similarities between the query frame and the database frames from \textbf{all testing scenarios in the dataset} are calculated and the database frames are ranked by the similarity. The Recall@k represents the percentage of matched frames in the top k candidates with the query frame. Besides Recall@k, we include other metrics such as running time (inference time), FLOPs, and model size as well. All those metrics are reported on the same PC equipped with one NVIDIA GeForce RTX 3060 GPU and an Intel Xeon(R) W-2255 CPU.

To evaluate the performance of our approach, PoCo, we compare with different SOTA indoor RGB-D and point cloud-based place recognition approaches, CGiS-Net~\cite{ming2022cgis}, MinkLoc-3D~\cite{komorowski2021minkloc3d}, NetVLAD~\cite{arandjelovic2016netvlad}, PointNet-VLAD~\cite{uy2018pointnetvlad}, Indoor DH3D~\cite{yang2022fd}. All these approaches are trained and tested in the two datasets, separately. 

For the ablation study, to demonstrate the capability of our model in handling both color and geometric information, we designed three types of experiments: 1. Vanilla RGB CoCs, which uses the CoC-Medium~\cite{cocs} as the backbone and applied our Encoder block of Fig.~\ref{fig:architecture}. This is to test the capability of CoCs in RGB place recognition, and also shows how much effect color information takes in place recognition task; 2. PoCo w/o Color, which only processes geometric information of the point cloud, where the input feature changes from $\set{r,g,b,x,y,z}$ to $\set{x,y,z}$. This ablation study tests the effectiveness of our PoCo model using geometric information in place recognition tasks; 3. PoCo w/o Eq. \ref{eq:geos}, which removes the relative geometric information and uses absolute positions of the points instead. This model is to test how much effect the relative geometric function has on the place recognition performance.

\begin{table}[h!]
    \centering
    \begin{tabular}{c|c|c|c}
ScanNetPR & R@1 $\uparrow$ &  R@2 $\uparrow$ &  R@3 $\uparrow$\\
\hline
SIFT~\cite{david2004distinctive} + BoW~\cite{sivic2008efficient} & 16.16 & 21.17 & 24.38 \\
\hline
NetVLAD~\cite{arandjelovic2016netvlad} & 21.77 & 33.81 & 41.49 \\
\hline
PointNet-VLAD~\cite{uy2018pointnetvlad} & 22.43 & 30.81 & 36.58 \\
\hline
Indoor DH3D~\cite{yang2022fd} & 16.10 & 21.92 & 25.30 \\
\hline
MinkLoc-3D~\cite{komorowski2021minkloc3d} & 10.13 & 16.63 & 20.80 \\
\hline
CGiS-Net~\cite{ming2022cgis} & 61.12 & 70.23 & 75.06 \\
\hline
CGiS-Net w/o color~\cite{ming2022cgis} & 39.62 & 50.92 & 56.14 \\
\hline
CGiS-Net w/o geometry~\cite{ming2022cgis} & 40.07 & 51.28 & 58.96 \\
\hline
\hline
Vanilla RGB CoCs~\cite{cocs} & 41.82 & 57.65 & 66.82 \\
\hline
PoCo w/o color & 44.34 & 54.27 & 59.78 \\
\hline
PoCo w/o Equation~\ref{eq:geos} & 62.23 & 73.81 & 79.63 \\
\hline
PoCo & \textbf{64.63} & \textbf{75.02} & \textbf{80.09} \\
    \end{tabular}
    \caption{The table shows our PoCo method outperforms other state-of-the-art methods by at least $5.7\%$ improvement in Recall@1 value in the ScanNetPR dataset.} 
    \label{tab:recall_scannet}
    \vspace{-1em}
\end{table}

\begin{table}[h!]
    \centering
    \begin{tabular}{c|c|c|c}
ARKit & R@1 $\uparrow$ &  R@2 $\uparrow$ &  R@3 $\uparrow$\\
\hline
PointNet-VLAD~\cite{uy2018pointnetvlad} & 11.04 & 16.57 & 20.57 \\
\hline
MinkLoc-3D~\cite{komorowski2021minkloc3d} & 8.14 & 10.95 & 13.79 \\
\hline
CGiS-Net~\cite{ming2022cgis} & 39.82 & 49.01 & 56.02 \\
\hline
\hline
Vanilla RGB CoCs~\cite{cocs} & 17.19 & 24.74 & 30.06 \\
\hline
PoCo w/o color & 21.58 & 30.00 & 35.66 \\
\hline
PoCo w/o Equation~\ref{eq:geos} & 41.41 & 51.86 & 58.11 \\
\hline
PoCo & \textbf{45.12} & \textbf{57.10} & \textbf{62.14} \\
    \end{tabular}
    \caption{The table shows our PoCo model outperforms other approaches by at least 4 points in Recall@1 in the ARKit dataset. 
    } 
    \label{tab:recall_arkit}
    \vspace{-1em}
\end{table}

\begin{table}[]
    \centering
    \begin{tabular}{c|c|c|c|c}
\multirow{ 2}{*}{Methods} & \multirow{ 2}{*}{R@1$\uparrow$} &  Running &  FLOPs & Model \\
 & &   Time (s) & (Mb)  &  Size (Mb)\\ \hline
 PointNet-VLAD & 22.43 & 0.01 & 1319.67 & 75.47 \\ \hline
 MinkLoc-3D & 10.13 &0.02 & - & 10.17 \\ \hline
 CGiS-Net & 61.12 & 0.07 & 738.60 & 26.18 \\ \hline
 PoCo & 64.63 & 0.04 & 285.63 & 29.94 \\ \hline
    \end{tabular}
    \caption{Our approach is $1.75$X faster than CGiS-Net and also has better recall values. Our computational cost is the smallest compared with other approaches.} 
    \label{tab:computation}
    \vspace{-2em}
\end{table}

\begin{figure*}[h!]
  \centering
  \begin{tabular}{ | c | c | c | c |c| }
    \hline
   \textbf{Query Frames} & \textbf{CGiS-Net Recall@1} & \textbf{PointNet-VLAD Recall@1} & \textbf{PoCo Recall@1} & \textbf{Ground Truth}  \\ \hline
   
    \begin{minipage}{.15\linewidth} \includegraphics[width=\linewidth,height=0.7\linewidth]{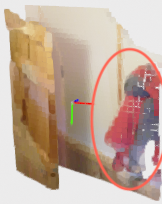} \end{minipage}
    &
    \begin{minipage}{.15\linewidth} \includegraphics[width=\linewidth,height=0.7\linewidth]{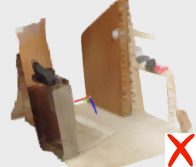} \end{minipage}
    &
    \begin{minipage}{.15\linewidth} \includegraphics[width=\linewidth,height=0.7\linewidth]{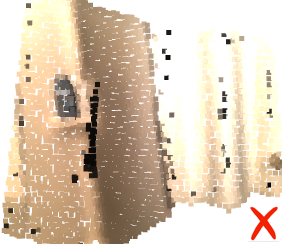} \end{minipage}
    &
    \begin{minipage}{.15\linewidth} \includegraphics[width=\linewidth,height=0.7\linewidth]{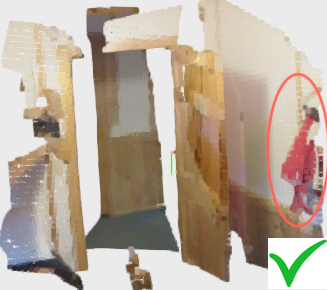} \end{minipage}
    &
    \begin{minipage}{.15\linewidth} \includegraphics[width=\linewidth,height=0.7\linewidth]{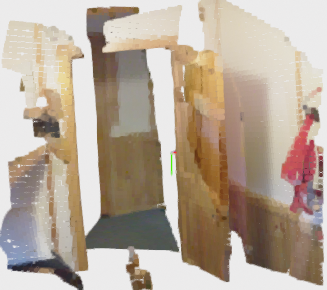} \end{minipage}
    \\ \hline
    
    \begin{minipage}{.15\linewidth} \includegraphics[width=\linewidth,height=0.7\linewidth]{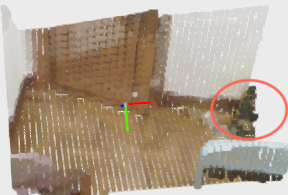} \end{minipage}
    &
    \begin{minipage}{.15\linewidth} \includegraphics[width=\linewidth,height=0.7\linewidth]{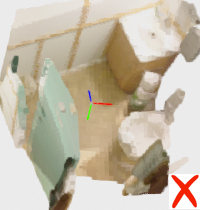} \end{minipage}
    &
    \begin{minipage}{.15\linewidth} \includegraphics[width=\linewidth,height=0.7\linewidth]{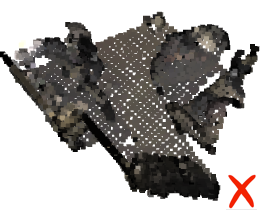} \end{minipage}
    &
    \begin{minipage}{.15\linewidth} \includegraphics[width=\linewidth,height=0.7\linewidth]{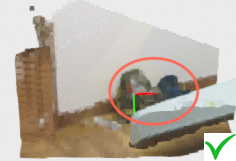} \end{minipage}
    &
    \begin{minipage}{.15\linewidth} \includegraphics[width=\linewidth,height=0.7\linewidth]{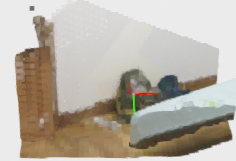} \end{minipage}
    \\ \hline
  \end{tabular}
  \caption{\textbf{Qualitative Comparison in the ScanNetPR dataset:} Our PoCo method outperforms other approaches in the challenging scenarios in the ScanNetPR dataset, with small overlap areas, marked by red circles, between query and positive frames.}
  \label{fig:qualitative_results_scannet}
  \vspace{-1em}
\end{figure*}

\begin{figure*}[h!]
  \centering
  \begin{tabular}{ | c | c | c | c |c| }
    \hline
   \textbf{Query Frames} & \textbf{CGiS-Net Recall@1} & \textbf{PointNet-VLAD Recall@1} & \textbf{PoCo Recall@1} & \textbf{Ground Truth}  \\ \hline
   
    \begin{minipage}{.15\linewidth} \includegraphics[width=\linewidth,height=0.7\linewidth]{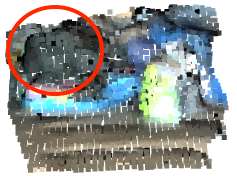} \end{minipage}
    &
    \begin{minipage}{.15\linewidth} \includegraphics[width=\linewidth,height=0.7\linewidth]{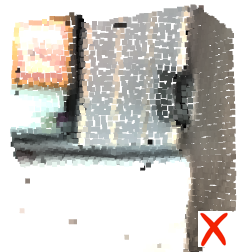} \end{minipage}
    &
    \begin{minipage}{.15\linewidth} \includegraphics[width=\linewidth,height=0.7\linewidth]{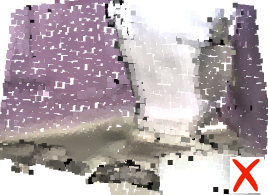} \end{minipage}
    &
    \begin{minipage}{.15\linewidth} \includegraphics[width=\linewidth,height=0.7\linewidth]{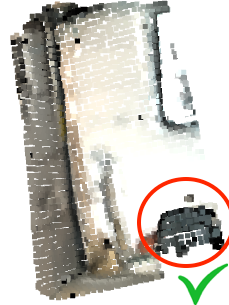} \end{minipage}
    &
    \begin{minipage}{.15\linewidth} \includegraphics[width=\linewidth,height=0.7\linewidth]{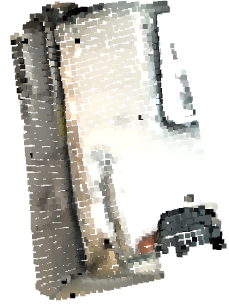} \end{minipage}
    \\ \hline
    
    \begin{minipage}{.15\linewidth} \includegraphics[width=\linewidth,height=0.7\linewidth]{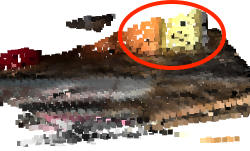} \end{minipage}
    &
    \begin{minipage}{.15\linewidth} \includegraphics[width=\linewidth,height=0.7\linewidth]{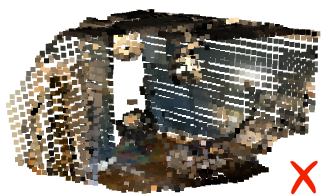} \end{minipage}
    &
    \begin{minipage}{.15\linewidth} \includegraphics[width=\linewidth,height=0.7\linewidth]{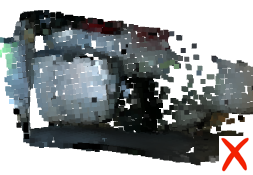} \end{minipage}
    &
    \begin{minipage}{.15\linewidth} \includegraphics[width=\linewidth,height=0.7\linewidth]{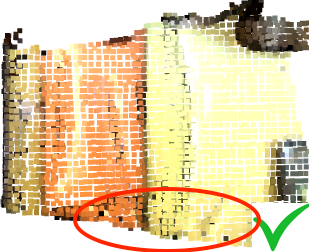} \end{minipage}
    &
    \begin{minipage}{.15\linewidth} \includegraphics[width=\linewidth,height=0.7\linewidth]{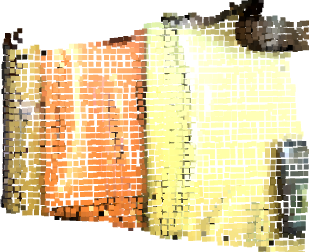} \end{minipage}
    \\ \hline
  \end{tabular}
  \caption{\textbf{Qualitative Comparison in the ARKit dataset:} Our PoCo method can detect small overlap areas, circled in red, but CGiS-Net and PointNet-VLAD fails.}
  \label{fig:qualitative_results_arkit}
  \vspace{-1em}
\end{figure*}

\begin{figure*}[h!]
  \centering
  \begin{tabular}{ | c | c | c | c| }
    \hline
   \textbf{Query Frames} & \textbf{PoCo w/o Color} & \textbf{PoCo} & \textbf{Ground Truth} \\ \hline
    \begin{minipage}{.22\textwidth}
    \includegraphics[width=\linewidth,height=0.7\linewidth]{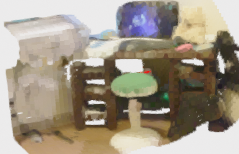}
    \end{minipage}
    &
    \begin{minipage}{.22\textwidth}
      \includegraphics[width=\linewidth,height=0.7\linewidth]{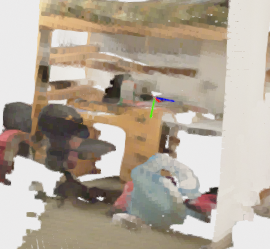}
    \end{minipage}
    & 
    \begin{minipage}{.22\textwidth}
      \includegraphics[width=\linewidth,height=0.7\linewidth]{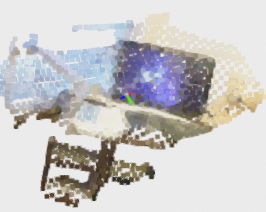}
    \end{minipage}
    & 
    \begin{minipage}{.22\textwidth}
      \includegraphics[width=\linewidth,height=0.7\linewidth]{figs/color/full0.png}
    \end{minipage}
    \\ \hline
    \begin{minipage}{.22\textwidth}
      \includegraphics[width=\linewidth,height=0.7\linewidth]{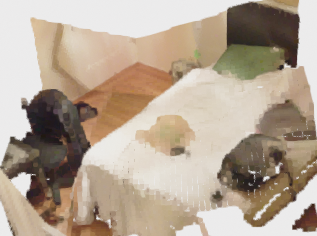}
    \end{minipage}
    &
    \begin{minipage}{.22\textwidth}
      \includegraphics[width=\linewidth,height=0.7\linewidth]{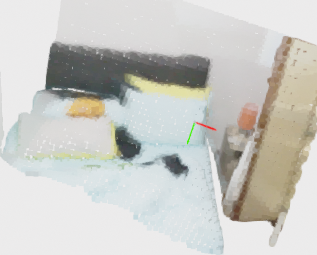}
    \end{minipage}
    & 
    \begin{minipage}{.22\textwidth}
      \includegraphics[width=\linewidth,height=0.7\linewidth]{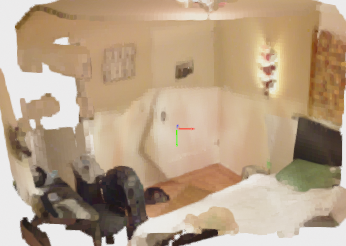}
    \end{minipage}
    & 
    \begin{minipage}{.22\textwidth}
      \includegraphics[width=\linewidth,height=0.7\linewidth]{figs/color/full_1.png}
    \end{minipage}
    \\ \hline
  \end{tabular}
  \caption{\textbf{Benefits of Color Information.} The second and third columns show recall@1 frame from models w/o and with color information, respectively. After we remove the color information, the PoCo w/o Color model selects candidate frames only based on geometric information, where the desk (1st row) and the bed (2nd row) in matched candidate frames are structurally similar to the query frames, but their colors are very different. }
  \label{fig:color_results}
  \vspace{-1.5em}
\end{figure*}

\begin{figure*}[h!]
  \centering
  \begin{tabular}{ | c | c | c | c| }
    \hline
   \textbf{Query Frames} & \textbf{Vanilla-RGB-PR} & \textbf{PoCo} & \textbf{Ground Truth} \\ \hline
    \begin{minipage}{.22\textwidth}
    \includegraphics[width=\linewidth,height=0.7\linewidth]{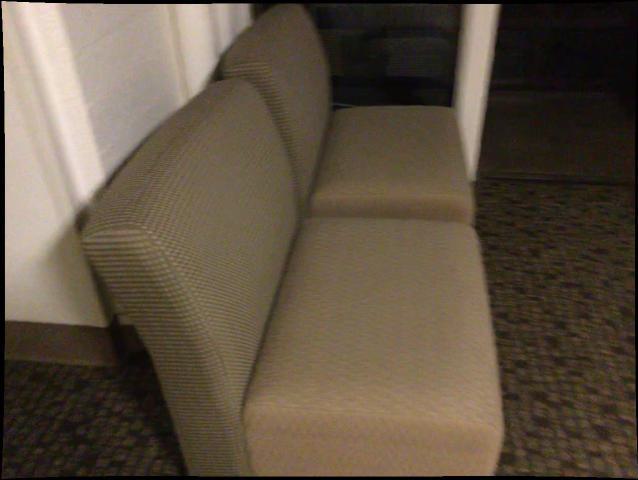}
    \end{minipage}
    &
    \begin{minipage}{.22\textwidth}
      \includegraphics[width=\linewidth,height=0.7\linewidth]{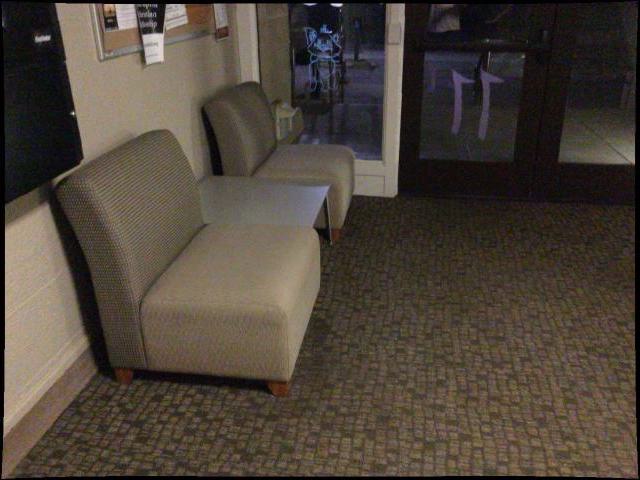}
    \end{minipage}
    & 
    \begin{minipage}{.22\textwidth}
      \includegraphics[width=\linewidth,height=0.7\linewidth]{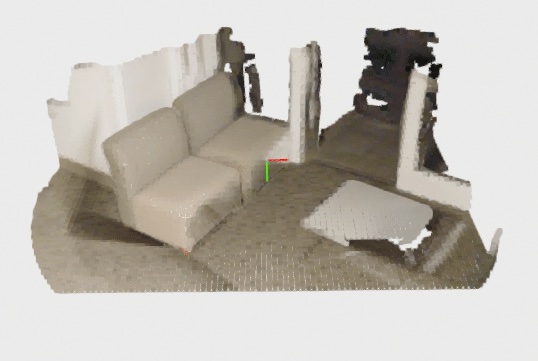}
    \end{minipage}
    & 
    \begin{minipage}{.22\textwidth}
      \includegraphics[width=\linewidth,height=0.7\linewidth]{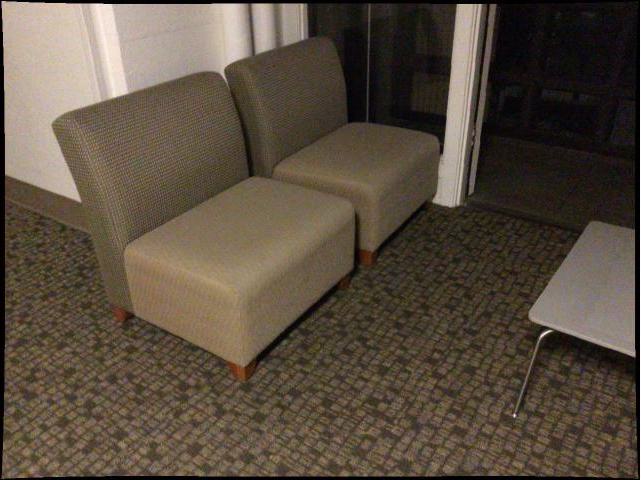}
    \end{minipage}
    \\ \hline
    \begin{minipage}{.22\textwidth}
      \includegraphics[width=\linewidth,height=0.7\linewidth]{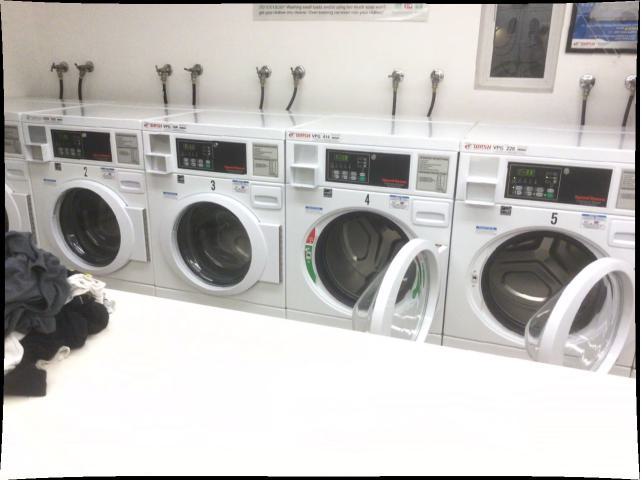}
    \end{minipage}
    &
    \begin{minipage}{.22\textwidth}
      \includegraphics[width=\linewidth,height=0.7\linewidth]{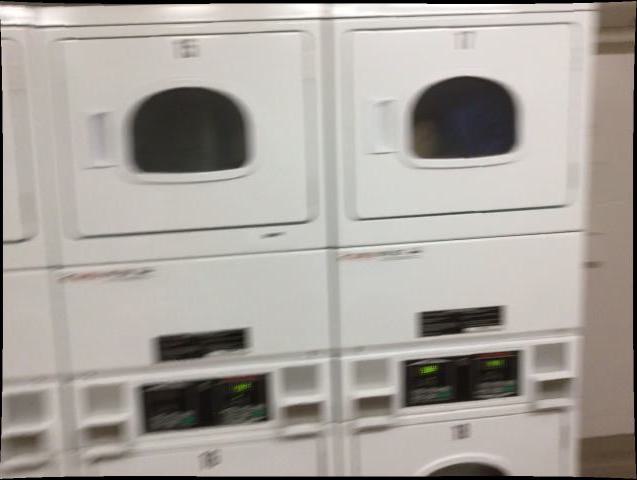}
    \end{minipage}
    & 
    \begin{minipage}{.22\textwidth}
      \includegraphics[width=\linewidth,height=0.7\linewidth]{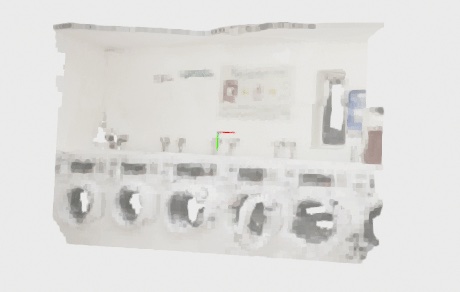}
    \end{minipage}
    & 
    \begin{minipage}{.22\textwidth}
      \includegraphics[width=\linewidth,height=0.7\linewidth]{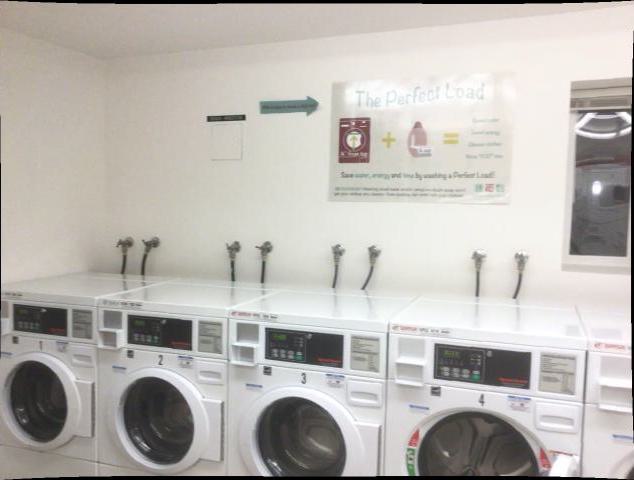}
    \end{minipage}
    \\ \hline
  \end{tabular}
  \caption{\textbf{Benefits of Geometric Information.} The Vanilla-RGB-PR can choose similar objects (e.g., chairs or washing machines). But it cannot capture spatial relationships of objects well. For example, in the second row, the query image has lined machines but  the matched image has stacked machines.}
  \label{fig:geometric_results}
  \vspace{-2em}
\end{figure*}

\subsection{Results}
In Table \ref{tab:recall_scannet} and Table \ref{tab:recall_arkit}, we present the performance of different SOTA approaches compared with our PoCo method and also ablation studies in two datasets. From Table \ref{tab:recall_scannet}, the learning-based methods show better accuracy in Recall@1 than SIFT+BoW. In the ScanNetPR dataset (Table \ref{tab:recall_scannet}), our PoCo method outperforms the other approaches by at least $5.7\%$  in Recall@1 and In the ARKit dataset (Table \ref{tab:recall_arkit}) we have at least $4$ points improvement. Considering PointNet-VLAD and MinkLoc-3D only process geometric information, in Table \ref{tab:recall_scannet}, we compare PoCo w/o color with these two methods and CGiS-Net w/o color. Our pure geometric model still outperforms other approaches by at least $11.91\%$ in Recall@1. In the ARKit dataset, our PoCo w/o color also outperforms other pure-geometric methods. Those results indicate our model can effectively encode geometric information for place recognition and support our innovation in generalizing CoCs to point cloud place recognition. As shown in Table~\ref{tab:recall_scannet}, compared with other RGB-based approaches NetVLAD~\cite{arandjelovic2016netvlad} and SIFT+BoW, our RGB model (Vanilla RGB CoCs) demonstrates better accuracy in Recall@1, which means our model is capable of effectively encoding the RGB information for place recognition.
From Fig.~\ref{fig:qualitative_results_scannet} and Fig.~\ref{fig:qualitative_results_arkit}, we observe our model is much better than baselines in those challenging cases, where the overlapping (red-circled) areas between query and positive frames are very small. From these two figures, we can also observe that PointNet-VLAD cannot recognize different colors for matching, whereas, in Fig.~\ref{fig:qualitative_results_scannet}, it chooses a scenario with a completely different color as the best match. Besides the accuracy, as shown in Table \ref{tab:computation}, we also compared the computational cost of different approaches. Compared with CGiS-Net, our PoCo model achieves both higher accuracy in Recall and less computational cost. The PointNet-VLAD and MinkLoc-3D have less inference time, but their accuracy is much worse than our model. 

About Ablation studies, as shown in Table \ref{tab:recall_scannet} and Table \ref{tab:recall_arkit}, geometric information and color information are both important to place recognition and they have very similar effects on the Recall values. Geometric information matters more than color information, and by only using each of them, in Table~\ref{tab:recall_scannet} PoCo w/o color has Recall@1 $2.52$ points higher than the RGB model. In Table~\ref{tab:recall_arkit}, it has a larger difference of $4.39$ points, which means that our design can effectively process geometric information for place recognition. Comparing Poco with PoCo w/o Equation~\ref{eq:geos}, we observe that relative geometric information makes the method more generalized to different point clouds and achieves 2.4 and 3.71 improvement in Recall@1 in ScanNetPR and ARKit datasets, respectively. Fig. \ref{fig:color_results} provides the qualitative difference between PoCo and RGB models. RGB model can only detect similar objects, but cannot tell complex positional relations among the object. Compared with it, our PoCo method can both recognize the objects and their relative positions in the scenarios. From Fig.~\ref{fig:geometric_results}, PoCo w/o Color is capable of recognizing very similar structures, such as the table and the chair in the first row. But because it doesn't have color information, it falsely matched the frames with very obvious color differences. However, PoCo is capable of utilizing color information to find the best match. From the ablation study, we demonstrate the efficacy of our innovation in jointly processing different modalities, color and geometric information, and achieves higher accuracy in indoor RGB-D place recognition tasks.

\section{Conclusion, Limitations and Future Work}
We are the first to propose an indoor RGB-D Point-Cloud-based place recognition model based on the concept of CoCs. We explicitly encode geometric information during feature extraction with transformer-based models. PoCo outperforms previous indoor RGB-D place recognition methods significantly in recall@K. However, our model still has its limitations, where sometimes our model relies much on geometric information which hurts the performance, and in some extreme cases, the overlapping areas between the ground truth and query frame are small and do not have many features. To solve the issues, we could use local features to help re-ranking candidate frames to improve accuracy. Therefore, in the future, a second stage of re-ranking candidates will be explored.

\noindent\textbf{Acknowledgment:} This work was supported in part by ARO Grants  W911NF2310046, W911NF2310352 and U.S. Army Cooperative Agreement W911NF2120076

\bibliographystyle{IEEEtran}
\bibliography{ref}


\end{document}